\documentclass{article}

     \usepackage[final, nonatbib]{neurips_2020}

\usepackage[utf8]{inputenc} %
\usepackage[T1]{fontenc}    %
\usepackage{hyperref}       %
\usepackage{url}            %
\usepackage{booktabs}       %
\usepackage{amsfonts}       %
\usepackage{nicefrac}       %
\usepackage{microtype}      %
\usepackage{epigraph}
\usepackage{wrapfig}

\usepackage{times}
\usepackage{epsfig}
\usepackage{amsmath}
\usepackage{amssymb}

\usepackage{xcolor} %

\usepackage{csquotes} %
\usepackage{mathtools} %
\usepackage{placeins} %
\usepackage{bbold}
\usepackage{breqn}
\usepackage{graphbox} %
\usepackage{multirow}

\usepackage[font=small]{caption}
\usepackage[font=small]{subcaption} %

\usepackage{url}
\usepackage{makecell}

\title{A Note on Data Biases in Generative Models}

\author{%

  Patrick Esser* \qquad\qquad Robin Rombach* \qquad\qquad Bj\"orn Ommer \\
  IWR, HCI, Heidelberg University \\
  \emph{firstname.lastname@iwr.uni-heidelberg.de}
}

\providecommand{\impath}[1]{}
\providecommand{\impatha}[1]{}
\providecommand{\impathb}[1]{}
\providecommand{\impathc}[1]{}
\providecommand{\imwidth}{}
\providecommand{\imwidtha}{}
\providecommand{\imwidthb}{}
\providecommand{\smallimwidtha}{}
\providecommand{\smallimwidthb}{}
\providecommand{\subimwidtha}{}
\providecommand{\subimwidthb}{}

\newcommand{\demo}{
 \begin{figure*}[hbtp]
   \renewcommand{\imwidth}{0.999\textwidth}
	\centering
 	\includegraphics[width=\imwidth]{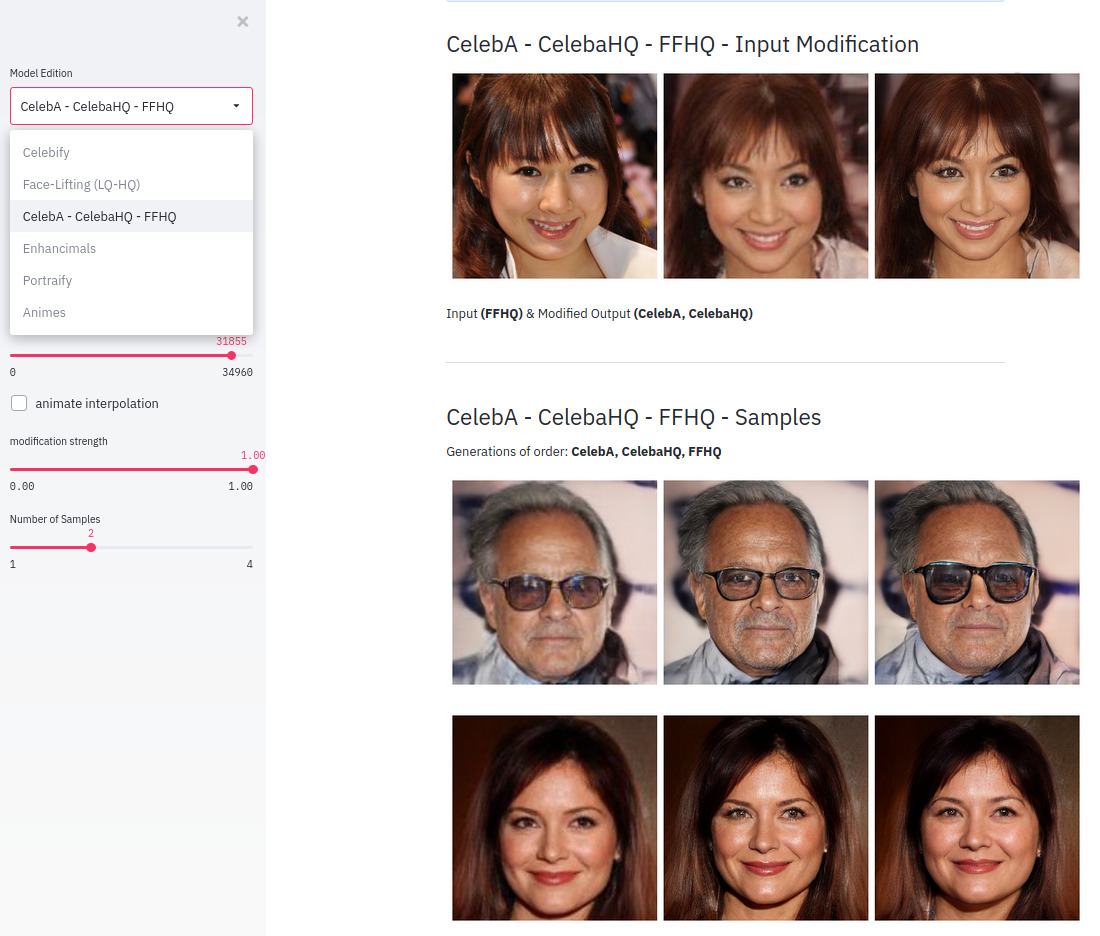}
 	\caption{Screenshot of our interactive web-based demonstration at \url{https://git.io/JkoKp}.}
 	\label{fig:demo}
 \end{figure*}
}

\newcommand{\progressastab}{
 \begin{figure*}[hbtp]
   \renewcommand{\imwidtha}{0.51\textwidth}
   \renewcommand{\imwidthb}{0.34\textwidth}
   \renewcommand{\smallimwidtha}{0.45\textwidth}
   \renewcommand{\smallimwidthb}{0.29\textwidth}
   \renewcommand{\subimwidtha}{0.165\textwidth}
   \renewcommand{\subimwidthb}{0.17\textwidth}
	\centering
	\begin{tabular}{c c}
    $\xrightarrow[\text{\parbox{\smallimwidtha}{\centering or progress on training data of faces?}}]{
      \text{\parbox{\smallimwidtha}{\centering progress on generative models of faces?}}}$ &
    $\xrightarrow[\text{\parbox{\smallimwidthb}{\centering or progress on animal data?}}]{\text{\parbox{\smallimwidthb}{\centering progress on animal
    models?}}}$ \\

 	\includegraphics[width=\imwidtha]{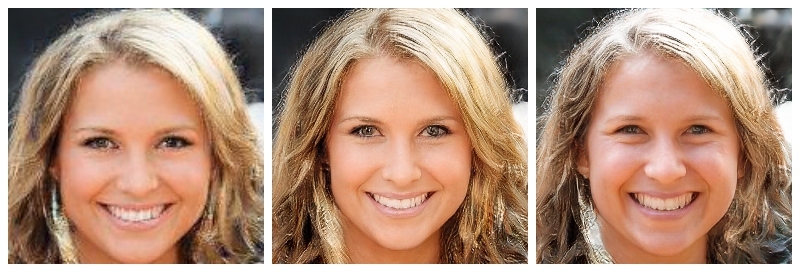} &
 	\includegraphics[width=\imwidthb]{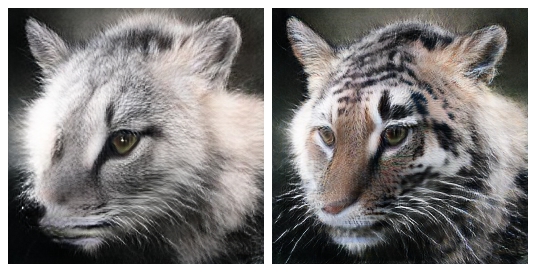} \\
 	\includegraphics[width=\imwidtha]{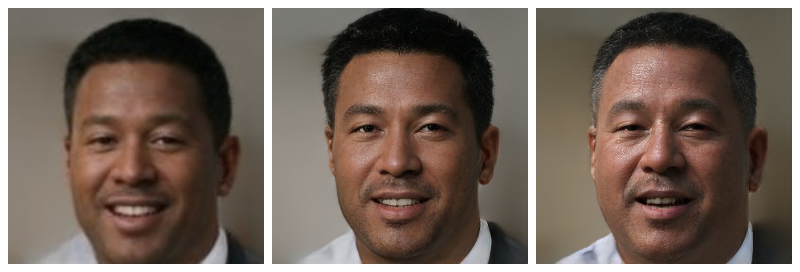} & 
 	\includegraphics[width=\imwidthb]{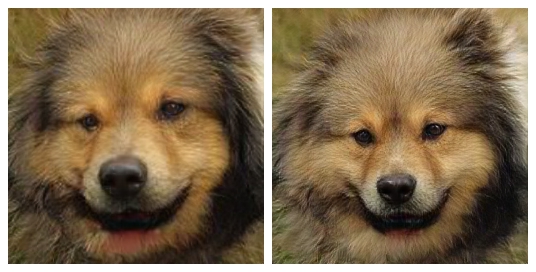} \\
    \parbox{\subimwidtha}{\centering \small 2015\\CelebA}
    \parbox{\subimwidtha}{\centering \small 2018\\CelebA-HQ}
    \parbox{\subimwidtha}{\centering \small 2019\\FFHQ} &
    \parbox{\subimwidthb}{\centering \small 2019\\AnimalFaces}
    \parbox{\subimwidthb}{\centering \small 2020\\AnimalFacesHQ}\\
	\end{tabular}
  \caption{Model or Data Progress? Our method can render \emph{the same}
   content projected onto \emph{different} datasets with \emph{the same} model, thereby questioning a popular
   conception on the effective progress of generative models.
   }
 	\label{fig:progress}
 \end{figure*}
}

\newcommand{\secondpagefigure}{
 \begin{figure*}[thbp]
   \renewcommand{\imwidth}{0.275\textwidth}
   \renewcommand{\impatha}[1]{img/portrait/##1}
   \renewcommand{\impathb}[1]{img/anime/##1}
   \renewcommand{\impathc}[1]{img/celebify/##1}
	\centering
	\begin{tabular}{c c c}
	\scriptsize{Oil-Portrait to Photography} &	\scriptsize{Anime to Photography} & \scriptsize{FFHQ to CelebA-HQ} \\
	\toprule
 	 	\vspace*{-0.2cm}
 	\includegraphics[width=\imwidth]{\impatha{newton}} & 
 	\includegraphics[width=\imwidth]{\impathb{15628_re}} & 
 	\includegraphics[width=\imwidth]{\impathc{10520}} \\
 	 	\vspace*{-0.2cm}
 	\includegraphics[width=\imwidth]{\impatha{mona2real}} &
 	\includegraphics[width=\imwidth]{\impathb{15633_re}} & 
 	\includegraphics[width=\imwidth]{\impathc{11347}} \\	
	 	\vspace*{-0.1cm} 	
 	\includegraphics[width=\imwidth]{\impatha{gal2real}}  & 	
 	\includegraphics[width=\imwidth]{\impathb{16368_re}} &
	\includegraphics[width=\imwidth]{\impathc{10618}}  \\

	\end{tabular}
  \setlength{\belowcaptionskip}{-1.5em}
 	\caption{Bringing oil portraits and animes to live by projecting them onto the FFHQ dataset (column 1 and 2, respectively). Column 3 visualizes the more subtle bias that is introduced when using different datasets of human faces. Here, we use our method to project images from the FFHQ dataset onto CelebA-HQ.}
 	\label{fig:secondpage}
 \end{figure*}
}

\newcommand{\anime}{
 \begin{figure*}[hbtp]
   \renewcommand{\imwidth}{0.3\textwidth}
   \renewcommand{\impath}[1]{img/anime/##1}
	\centering
 	\includegraphics[width=\imwidth]{\impath{1}}
 	\includegraphics[width=\imwidth]{\impath{5645}}
 	\includegraphics[width=\imwidth]{\impath{10420}}
 	\includegraphics[width=\imwidth]{\impath{10584}}
 	\includegraphics[width=\imwidth]{\impath{11181}}
 	\includegraphics[width=\imwidth]{\impath{11236}}
 	\caption{Image-to-Anime Transfer.}
 	\label{fig:anime}
 \end{figure*}
}

\newcommand{\celebifyinputs}{
 \begin{figure*}[hbtp]
   \renewcommand{\imwidth}{0.3\textwidth}
   \renewcommand{\impath}[1]{img/celebify/##1}
	\centering
 	\includegraphics[width=\imwidth]{\impath{9966}}
 	\includegraphics[width=\imwidth]{\impath{10945}}
 	\includegraphics[width=\imwidth]{\impath{10956}}
 	\includegraphics[width=\imwidth]{\impath{10997}}
 	\includegraphics[width=\imwidth]{\impath{12186}}
 	\includegraphics[width=\imwidth]{\impath{12531}}
 	\caption{FFHQ to CelebA-HQ Transfer.}
 	\label{fig:celebifyinputs}
 \end{figure*}
}

\newcommand{\portraits}{
 \begin{figure*}[hbtp]
   \renewcommand{\imwidth}{0.3\textwidth}
   \renewcommand{\impath}[1]{img/portrait/##1}
	\centering
 	\includegraphics[width=\imwidth]{\impath{coriolis2real}}
 	\includegraphics[width=\imwidth]{\impath{gauss2real}}
 	\includegraphics[width=\imwidth]{\impath{kant2real}}
 	\includegraphics[width=\imwidth]{\impath{mozart}}
 	\includegraphics[width=\imwidth]{\impath{rousseau}}
 	\includegraphics[width=\imwidth]{\impath{p2r_1}}
 	\includegraphics[width=\imwidth]{\impath{sample1}}
 	\includegraphics[width=\imwidth]{\impath{sample2}}
  	\includegraphics[width=\imwidth]{\impath{sample3}}
 	\caption{Portrait-to-Image transfer (row 1, 2) and samples (row 3), all
   synthesized with the same model.}
 	\label{fig:portraits}
 \end{figure*}
}

\newcommand{\animalsamples}{
\begin{figure}[htbp]
  \renewcommand{\impath}[1]{img/facelift_samples/enhancimals/##1}
  \renewcommand{\imwidth}{0.12\textwidth}
  \setlength{\tabcolsep}{2pt}
  \centering
  \begin{tabular}{ c  c  c  c  c  c  c  c  c}
    \multicolumn{4}{c}{\emph{AnimalFacesHQ}} & \phantom{x} & \multicolumn{4}{c}{\emph{AnimalFaces}}\\
    \toprule
	\includegraphics[width=\imwidth, align=c]{\impath{hq-0001}} &     
	\includegraphics[width=\imwidth, align=c]{\impath{hq-0002}} &         
	\includegraphics[width=\imwidth, align=c]{\impath{hq-0003}} &         
	\includegraphics[width=\imwidth, align=c]{\impath{hq-0004}} &         
	\phantom{x} & 
	\includegraphics[width=\imwidth, align=c]{\impath{lq-0001}} &         
	\includegraphics[width=\imwidth, align=c]{\impath{lq-0002}} &         
    \includegraphics[width=\imwidth, align=c]{\impath{lq-0003}} &         
	\includegraphics[width=\imwidth, align=c]{\impath{lq-0004}} \\   
	\midrule
  	\includegraphics[width=\imwidth, align=c]{\impath{hq-0005}} &     
	\includegraphics[width=\imwidth, align=c]{\impath{hq-0006}} &         
	\includegraphics[width=\imwidth, align=c]{\impath{hq-0007}} &         
	\includegraphics[width=\imwidth, align=c]{\impath{hq-0008}} &         
	\phantom{x} & 
	\includegraphics[width=\imwidth, align=c]{\impath{lq-0005}} &         
	\includegraphics[width=\imwidth, align=c]{\impath{lq-0006}} &         
    \includegraphics[width=\imwidth, align=c]{\impath{lq-0007}} &         
	\includegraphics[width=\imwidth, align=c]{\impath{lq-0008}} \\   
	\midrule
  	\includegraphics[width=\imwidth, align=c]{\impath{hq-0011}} &     
	\includegraphics[width=\imwidth, align=c]{\impath{hq-0012}} &         
	\includegraphics[width=\imwidth, align=c]{\impath{hq-0013}} &         
	\includegraphics[width=\imwidth, align=c]{\impath{hq-0014}} &         
	\phantom{x} & 
	\includegraphics[width=\imwidth, align=c]{\impath{lq-0011}} &         
	\includegraphics[width=\imwidth, align=c]{\impath{lq-0012}} &         
    \includegraphics[width=\imwidth, align=c]{\impath{lq-0013}} &         
	\includegraphics[width=\imwidth, align=c]{\impath{lq-0014}} \\   
	\midrule
  	\includegraphics[width=\imwidth, align=c]{\impath{hq-0015}} &     
	\includegraphics[width=\imwidth, align=c]{\impath{hq-0016}} &         
	\includegraphics[width=\imwidth, align=c]{\impath{hq-0017}} &         
	\includegraphics[width=\imwidth, align=c]{\impath{hq-0018}} &         
	\phantom{x} & 
	\includegraphics[width=\imwidth, align=c]{\impath{lq-0015}} &         
	\includegraphics[width=\imwidth, align=c]{\impath{lq-0016}} &         
    \includegraphics[width=\imwidth, align=c]{\impath{lq-0017}} &         
	\includegraphics[width=\imwidth, align=c]{\impath{lq-0018}} \\   
    \bottomrule  
  \end{tabular}    
  \caption{Samples from our model, conditioned on AnimalFacesHQ (left) and the older AnimalFaces (right).}
  \label{fig:animalsamples}
\end{figure}
}

\newcommand{\faceliftsamples}{
\begin{figure}[htbp]
  \renewcommand{\impath}[1]{img/facelift_samples/facelift/##1}
  \renewcommand{\imwidth}{0.12\textwidth}
  \setlength{\tabcolsep}{2pt}
  \centering
  \begin{tabular}{ c  c  c  c  c  c  c  c  c}
    \multicolumn{4}{c}{\emph{FacesHQ}} & \phantom{x} & \multicolumn{4}{c}{\emph{CelebA}}\\
    \toprule
	\includegraphics[width=\imwidth, align=c]{\impath{hq-0001}} &     
	\includegraphics[width=\imwidth, align=c]{\impath{hq-0002}} &         
	\includegraphics[width=\imwidth, align=c]{\impath{hq-0003}} &         
	\includegraphics[width=\imwidth, align=c]{\impath{hq-0004}} &         
	\phantom{x} & 
	\includegraphics[width=\imwidth, align=c]{\impath{lq-0001}} &         
	\includegraphics[width=\imwidth, align=c]{\impath{lq-0002}} &         
    \includegraphics[width=\imwidth, align=c]{\impath{lq-0003}} &         
	\includegraphics[width=\imwidth, align=c]{\impath{lq-0004}} \\   
	\midrule
  	\includegraphics[width=\imwidth, align=c]{\impath{hq-0005}} &     
	\includegraphics[width=\imwidth, align=c]{\impath{hq-0006}} &         
	\includegraphics[width=\imwidth, align=c]{\impath{hq-0007}} &         
	\includegraphics[width=\imwidth, align=c]{\impath{hq-0008}} &         
	\phantom{x} & 
	\includegraphics[width=\imwidth, align=c]{\impath{lq-0005}} &         
	\includegraphics[width=\imwidth, align=c]{\impath{lq-0006}} &         
    \includegraphics[width=\imwidth, align=c]{\impath{lq-0007}} &         
	\includegraphics[width=\imwidth, align=c]{\impath{lq-0008}} \\   
	\midrule
  	\includegraphics[width=\imwidth, align=c]{\impath{hq-0011}} &     
	\includegraphics[width=\imwidth, align=c]{\impath{hq-0012}} &         
	\includegraphics[width=\imwidth, align=c]{\impath{hq-0013}} &         
	\includegraphics[width=\imwidth, align=c]{\impath{hq-0014}} &         
	\phantom{x} & 
	\includegraphics[width=\imwidth, align=c]{\impath{lq-0011}} &         
	\includegraphics[width=\imwidth, align=c]{\impath{lq-0012}} &         
    \includegraphics[width=\imwidth, align=c]{\impath{lq-0013}} &         
	\includegraphics[width=\imwidth, align=c]{\impath{lq-0014}} \\   
	\midrule
  	\includegraphics[width=\imwidth, align=c]{\impath{hq-0015}} &     
	\includegraphics[width=\imwidth, align=c]{\impath{hq-0016}} &         
	\includegraphics[width=\imwidth, align=c]{\impath{hq-0017}} &         
	\includegraphics[width=\imwidth, align=c]{\impath{hq-0018}} &         
	\phantom{x} & 
	\includegraphics[width=\imwidth, align=c]{\impath{lq-0015}} &         
	\includegraphics[width=\imwidth, align=c]{\impath{lq-0016}} &         
    \includegraphics[width=\imwidth, align=c]{\impath{lq-0017}} &         
	\includegraphics[width=\imwidth, align=c]{\impath{lq-0018}} \\   
    \bottomrule  
  \end{tabular}    
  \caption{Samples from our model, conditioned on the concatenated FacesHQ
  dataset (consisting of CelebA-HQ and FFHQ) (left column) and the CelebA dataset (right column).}
  \label{fig:faceliftsamples}
\end{figure}
}

\newcommand{\celebifysamples}{
\begin{figure}[htbp]
  \renewcommand{\impath}[1]{img/facelift_samples/celebify/##1}
  \renewcommand{\imwidth}{0.12\textwidth}
  \setlength{\tabcolsep}{2pt}
  \centering
  \begin{tabular}{ c  c  c  c  c  c  c  c  c}
    \multicolumn{4}{c}{\emph{FFHQ}} & \phantom{x} & \multicolumn{4}{c}{\emph{CelebA-HQ}}\\
    \toprule
	\includegraphics[width=\imwidth, align=c]{\impath{hq-0001}} &     
	\includegraphics[width=\imwidth, align=c]{\impath{hq-0002}} &         
	\includegraphics[width=\imwidth, align=c]{\impath{hq-0003}} &         
	\includegraphics[width=\imwidth, align=c]{\impath{hq-0004}} &         
	\phantom{x} & 
	\includegraphics[width=\imwidth, align=c]{\impath{lq-0001}} &         
	\includegraphics[width=\imwidth, align=c]{\impath{lq-0002}} &         
    \includegraphics[width=\imwidth, align=c]{\impath{lq-0003}} &         
	\includegraphics[width=\imwidth, align=c]{\impath{lq-0004}} \\   
	\midrule
  	\includegraphics[width=\imwidth, align=c]{\impath{hq-0009}} &     
	\includegraphics[width=\imwidth, align=c]{\impath{hq-0006}} &         
	\includegraphics[width=\imwidth, align=c]{\impath{hq-0007}} &         
	\includegraphics[width=\imwidth, align=c]{\impath{hq-0008}} &         
	\phantom{x} & 
	\includegraphics[width=\imwidth, align=c]{\impath{lq-0009}} &         
	\includegraphics[width=\imwidth, align=c]{\impath{lq-0006}} &         
    \includegraphics[width=\imwidth, align=c]{\impath{lq-0007}} &         
	\includegraphics[width=\imwidth, align=c]{\impath{lq-0008}} \\   
	\midrule
  	\includegraphics[width=\imwidth, align=c]{\impath{hq-0011}} &     
	\includegraphics[width=\imwidth, align=c]{\impath{hq-0010}} &         
	\includegraphics[width=\imwidth, align=c]{\impath{hq-0013}} &         
	\includegraphics[width=\imwidth, align=c]{\impath{hq-0014}} &         
	\phantom{x} & 
	\includegraphics[width=\imwidth, align=c]{\impath{lq-0011}} &         
	\includegraphics[width=\imwidth, align=c]{\impath{lq-0010}} &         
    \includegraphics[width=\imwidth, align=c]{\impath{lq-0013}} &         
	\includegraphics[width=\imwidth, align=c]{\impath{lq-0014}} \\   
	\midrule
  	\includegraphics[width=\imwidth, align=c]{\impath{hq-0015}} &     
	\includegraphics[width=\imwidth, align=c]{\impath{hq-0016}} &         
	\includegraphics[width=\imwidth, align=c]{\impath{hq-0017}} &         
	\includegraphics[width=\imwidth, align=c]{\impath{hq-0018}} &         
	\phantom{x} & 
	\includegraphics[width=\imwidth, align=c]{\impath{lq-0015}} &         
	\includegraphics[width=\imwidth, align=c]{\impath{lq-0016}} &         
    \includegraphics[width=\imwidth, align=c]{\impath{lq-0017}} &         
	\includegraphics[width=\imwidth, align=c]{\impath{lq-0018}} \\   
    \bottomrule  
  \end{tabular}    
  \caption{Samples from our model, conditioned on the FFHQ dataset (left) and the CelebA-HQ dataset (right).}
  \label{fig:celebifysamples}
\end{figure}
}

\newcommand{\animesamples}{
\begin{figure}[htbp]
  \renewcommand{\impath}[1]{img/facelift_samples/animes/##1}
  \renewcommand{\imwidth}{0.12\textwidth}
  \setlength{\tabcolsep}{2pt}
  \centering
  \begin{tabular}{ c  c  c  c  c  c  c  c  c}
    \multicolumn{4}{c}{\emph{Anime}} & \phantom{x} & \multicolumn{4}{c}{\emph{Photography}}\\
    \toprule
	\includegraphics[width=\imwidth, align=c]{\impath{hq-0001}} &     
	\includegraphics[width=\imwidth, align=c]{\impath{hq-0002}} &         
	\includegraphics[width=\imwidth, align=c]{\impath{hq-0003}} &         
	\includegraphics[width=\imwidth, align=c]{\impath{hq-0004}} &         
	\phantom{x} & 
	\includegraphics[width=\imwidth, align=c]{\impath{lq-0001}} &         
	\includegraphics[width=\imwidth, align=c]{\impath{lq-0002}} &         
    \includegraphics[width=\imwidth, align=c]{\impath{lq-0003}} &         
	\includegraphics[width=\imwidth, align=c]{\impath{lq-0004}} \\   
	\midrule
  	\includegraphics[width=\imwidth, align=c]{\impath{hq-0009}} &     
	\includegraphics[width=\imwidth, align=c]{\impath{hq-0006}} &         
	\includegraphics[width=\imwidth, align=c]{\impath{hq-0007}} &         
	\includegraphics[width=\imwidth, align=c]{\impath{hq-0008}} &         
	\phantom{x} & 
	\includegraphics[width=\imwidth, align=c]{\impath{lq-0009}} &         
	\includegraphics[width=\imwidth, align=c]{\impath{lq-0006}} &         
    \includegraphics[width=\imwidth, align=c]{\impath{lq-0007}} &         
	\includegraphics[width=\imwidth, align=c]{\impath{lq-0008}} \\   
	\midrule
  	\includegraphics[width=\imwidth, align=c]{\impath{hq-0019}} &     
	\includegraphics[width=\imwidth, align=c]{\impath{hq-0010}} &         
	\includegraphics[width=\imwidth, align=c]{\impath{hq-0013}} &         
	\includegraphics[width=\imwidth, align=c]{\impath{hq-0020}} &         
	\phantom{x} & 
	\includegraphics[width=\imwidth, align=c]{\impath{lq-0019}} &         
	\includegraphics[width=\imwidth, align=c]{\impath{lq-0010}} &         
    \includegraphics[width=\imwidth, align=c]{\impath{lq-0013}} &         
	\includegraphics[width=\imwidth, align=c]{\impath{lq-0020}} \\   
	\midrule
  	\includegraphics[width=\imwidth, align=c]{\impath{hq-0015}} &     
	\includegraphics[width=\imwidth, align=c]{\impath{hq-0016}} &         
	\includegraphics[width=\imwidth, align=c]{\impath{hq-0017}} &         
	\includegraphics[width=\imwidth, align=c]{\impath{hq-0018}} &         
	\phantom{x} & 
	\includegraphics[width=\imwidth, align=c]{\impath{lq-0015}} &         
	\includegraphics[width=\imwidth, align=c]{\impath{lq-0016}} &         
    \includegraphics[width=\imwidth, align=c]{\impath{lq-0017}} &         
	\includegraphics[width=\imwidth, align=c]{\impath{lq-0018}} \\   
    \bottomrule  
  \end{tabular}    
  \caption{Samples from our model, conditioned on the Anime dataset (left) and the FFHQ dataset (right).}
  \label{fig:animesamples}
\end{figure}
}

\newcommand{\portraitsamples}{
\begin{figure}[htbp]
  \renewcommand{\impath}[1]{img/facelift_samples/portraits/##1}
  \renewcommand{\imwidth}{0.12\textwidth}
  \setlength{\tabcolsep}{2pt}
  \centering
  \begin{tabular}{ c  c  c  c  c  c  c  c  c}
    \multicolumn{4}{c}{\emph{Oil Portraits}} & \phantom{x} & \multicolumn{4}{c}{\emph{Photography}}\\
    \toprule
	\includegraphics[width=\imwidth, align=c]{\impath{hq-0001}} &     
	\includegraphics[width=\imwidth, align=c]{\impath{hq-0002}} &         
	\includegraphics[width=\imwidth, align=c]{\impath{hq-0003}} &         
	\includegraphics[width=\imwidth, align=c]{\impath{hq-0004}} &         
	\phantom{x} & 
	\includegraphics[width=\imwidth, align=c]{\impath{lq-0001}} &         
	\includegraphics[width=\imwidth, align=c]{\impath{lq-0002}} &         
    \includegraphics[width=\imwidth, align=c]{\impath{lq-0003}} &         
	\includegraphics[width=\imwidth, align=c]{\impath{lq-0004}} \\   
	\midrule
  	\includegraphics[width=\imwidth, align=c]{\impath{hq-0009}} &     
	\includegraphics[width=\imwidth, align=c]{\impath{hq-0006}} &         
	\includegraphics[width=\imwidth, align=c]{\impath{hq-0007}} &         
	\includegraphics[width=\imwidth, align=c]{\impath{hq-0008}} &         
	\phantom{x} & 
	\includegraphics[width=\imwidth, align=c]{\impath{lq-0009}} &         
	\includegraphics[width=\imwidth, align=c]{\impath{lq-0006}} &         
    \includegraphics[width=\imwidth, align=c]{\impath{lq-0007}} &         
	\includegraphics[width=\imwidth, align=c]{\impath{lq-0008}} \\   
	\midrule
  	\includegraphics[width=\imwidth, align=c]{\impath{hq-0019}} &     
	\includegraphics[width=\imwidth, align=c]{\impath{hq-0010}} &         
	\includegraphics[width=\imwidth, align=c]{\impath{hq-0013}} &         
	\includegraphics[width=\imwidth, align=c]{\impath{hq-0020}} &         
	\phantom{x} & 
	\includegraphics[width=\imwidth, align=c]{\impath{lq-0019}} &         
	\includegraphics[width=\imwidth, align=c]{\impath{lq-0010}} &         
    \includegraphics[width=\imwidth, align=c]{\impath{lq-0013}} &         
	\includegraphics[width=\imwidth, align=c]{\impath{lq-0020}} \\   
	\midrule
  	\includegraphics[width=\imwidth, align=c]{\impath{hq-0015}} &     
	\includegraphics[width=\imwidth, align=c]{\impath{hq-0016}} &         
	\includegraphics[width=\imwidth, align=c]{\impath{hq-0017}} &         
	\includegraphics[width=\imwidth, align=c]{\impath{hq-0018}} &         
	\phantom{x} & 
	\includegraphics[width=\imwidth, align=c]{\impath{lq-0015}} &         
	\includegraphics[width=\imwidth, align=c]{\impath{lq-0016}} &         
    \includegraphics[width=\imwidth, align=c]{\impath{lq-0017}} &         
	\includegraphics[width=\imwidth, align=c]{\impath{lq-0018}} \\   
    \bottomrule  
  \end{tabular}    
  \caption{Samples from our model, conditioned on the Portrait dataset (left) and the FFHQ dataset (right).}
  \label{fig:portraitsamples}
\end{figure}
}

\newcommand{\triplesamples}{
\begin{figure}[htbp]
  \renewcommand{\impath}[1]{img/facelift_samples/triplefaces/##1}
  \renewcommand{\imwidth}{0.1025\textwidth}
  \setlength{\tabcolsep}{2pt}
  \centering
  \begin{tabular}{ c  c  c    c    c  c  c    c   c  c  c}
    \multicolumn{3}{c}{\emph{CelebA}} & \phantom{x} & \multicolumn{3}{c}{\emph{CelebA-HQ}} & \phantom{x} & \multicolumn{3}{c}{\emph{FFHQ}} \\
    \toprule
	\includegraphics[width=\imwidth, align=c]{\impath{CelebA-0001}} &     
	\includegraphics[width=\imwidth, align=c]{\impath{CelebA-0002}} &         
	\includegraphics[width=\imwidth, align=c]{\impath{CelebA-0003}} &         
	\phantom{x} & 
	\includegraphics[width=\imwidth, align=c]{\impath{CelebaHQ-0001}} &         
	\includegraphics[width=\imwidth, align=c]{\impath{CelebaHQ-0002}} &         
	\includegraphics[width=\imwidth, align=c]{\impath{CelebaHQ-0003}} &
	\phantom{x} & 
	\includegraphics[width=\imwidth, align=c]{\impath{FFHQ-0001}} &         
	\includegraphics[width=\imwidth, align=c]{\impath{FFHQ-0002}} &         
	\includegraphics[width=\imwidth, align=c]{\impath{FFHQ-0003}} \\
	\midrule
	\includegraphics[width=\imwidth, align=c]{\impath{CelebA-0004}} &     
	\includegraphics[width=\imwidth, align=c]{\impath{CelebA-0005}} &         
	\includegraphics[width=\imwidth, align=c]{\impath{CelebA-0006}} &         
	\phantom{x} & 
	\includegraphics[width=\imwidth, align=c]{\impath{CelebaHQ-0004}} &         
	\includegraphics[width=\imwidth, align=c]{\impath{CelebaHQ-0005}} &         
	\includegraphics[width=\imwidth, align=c]{\impath{CelebaHQ-0006}} &
	\phantom{x} & 
	\includegraphics[width=\imwidth, align=c]{\impath{FFHQ-0004}} &         
	\includegraphics[width=\imwidth, align=c]{\impath{FFHQ-0005}} &         
	\includegraphics[width=\imwidth, align=c]{\impath{FFHQ-0006}} \\
	\midrule
	\includegraphics[width=\imwidth, align=c]{\impath{CelebA-0007}} &     
	\includegraphics[width=\imwidth, align=c]{\impath{CelebA-0008}} &         
	\includegraphics[width=\imwidth, align=c]{\impath{CelebA-0009}} &         
	\phantom{x} & 
	\includegraphics[width=\imwidth, align=c]{\impath{CelebaHQ-0007}} &         
	\includegraphics[width=\imwidth, align=c]{\impath{CelebaHQ-0008}} &         
	\includegraphics[width=\imwidth, align=c]{\impath{CelebaHQ-0009}} &
	\phantom{x} & 
	\includegraphics[width=\imwidth, align=c]{\impath{FFHQ-0007}} &         
	\includegraphics[width=\imwidth, align=c]{\impath{FFHQ-0008}} &         
	\includegraphics[width=\imwidth, align=c]{\impath{FFHQ-0009}} \\
	\midrule
	\includegraphics[width=\imwidth, align=c]{\impath{CelebA-0011}} &     
	\includegraphics[width=\imwidth, align=c]{\impath{CelebA-0012}} &         
	\includegraphics[width=\imwidth, align=c]{\impath{CelebA-0013}} &         
	\phantom{x} & 
	\includegraphics[width=\imwidth, align=c]{\impath{CelebaHQ-0011}} &         
	\includegraphics[width=\imwidth, align=c]{\impath{CelebaHQ-0012}} &         
	\includegraphics[width=\imwidth, align=c]{\impath{CelebaHQ-0013}} &
	\phantom{x} & 
	\includegraphics[width=\imwidth, align=c]{\impath{FFHQ-0011}} &         
	\includegraphics[width=\imwidth, align=c]{\impath{FFHQ-0012}} &         
	\includegraphics[width=\imwidth, align=c]{\impath{FFHQ-0013}} \\
    \bottomrule  
  \end{tabular}    
  \caption{Samples from our model, conditioned on the CelebA, CelebA-HQ and FFHQ datasets (left to right).}
  \label{fig:triplesamples}
\end{figure}
}

\newcommand{\figganprogress}{
\begin{figure}[htbp]
  \centering
	\includegraphics[width=\linewidth, align=c]{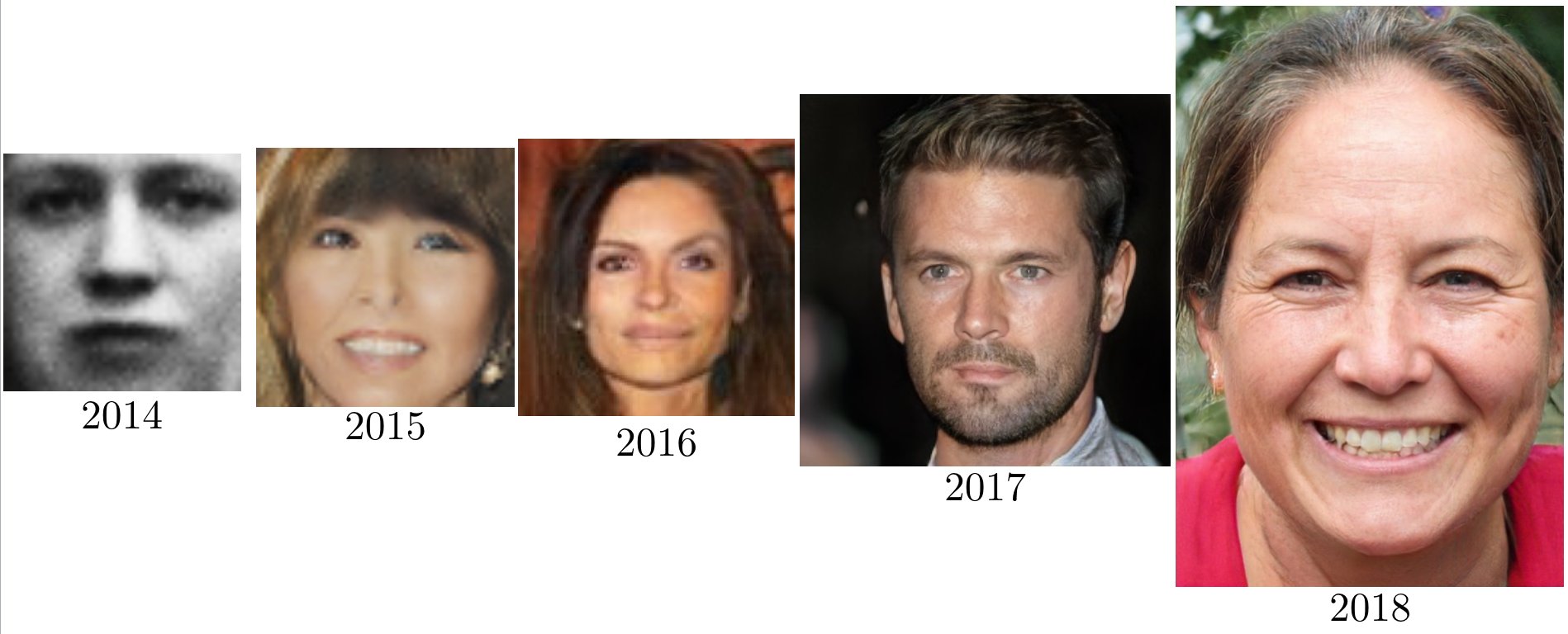}
  \caption{Reproduced from \cite{ganprogress}, where it was used to illustrate
  ``4.5 years of GAN progress''. The three rightmost figures were obtained with
  three different models on three different datasets. In contrast, Fig.~\ref{fig:progress}
  shows samples from \emph{a single} model on the same three datasets.}
  \label{fig:ganprogress}
\end{figure}
}

\begin{document}

\maketitle
\vspace{-1.5em}

\progressastab

\paragraph{Awareness for Biases}
\begin{wrapfigure}{r}{0pt}
    \begin{minipage}{0.43\linewidth} %
  \vspace{-1.8em}
\epigraph{\textit{``ML systems are biased when data is biased.''}}{---Yann LeCun \cite{lecunbias}}
  \vspace{-1.8em}
    \end{minipage}
\end{wrapfigure}

\vspace{-1.4em}
It is tempting to think that machines are less prone to unfairness and
prejudice. However, machine learning (ML) approaches compute their outputs based on
data. While biases can enter at any stage of the ML development pipeline
\cite{gupta2020state}, models are particularly receptive to mirror biases
of the datasets they are trained on.
Because the quality of training datasets play an important role in the quality
of trained models, practitioners are typically well aware that predictions of
ML systems do not necessarily reflect truths about the world but, first of all,
truths about the data \cite{Torralba2011UnbiasedLA}. However, this knowledge
can also be misused to conceal agendas behind a cloak of objectivity
\cite{kilchertrash}.
It is therefore increasingly important to raise awareness about the
relationship between modern algorithms and the data that shape them.
Besides the fact that art in general, and visual illustrations (e.g.\ images) of
this interplay in particular are already a good tool to make this relationship
accessible to a wider, non-specialist, audience, the value
of \emph{interactive} demonstrations that offer the possibility to analyze your
own data cannot be underestimated \cite{kurenkov2020lessons}.
Thus, we provide code for a web-based demonstration at \url{https://git.io/JkoKp}
(see Fig.~\ref{fig:demo}), which allows for an interactive exploration of the effects when models are
trained on biased data.%

\vspace{-0.8em}
\paragraph{Disentangling Dataset and Content}
Our goal is to demonstrate how the choice of a particular dataset affects
the outcome of a generative model.
We aim to project an image onto different datasets, i.e.\ visualize how this
image looks in a data-biased model.
Thus, we need to recover the generating factor which is independent of
the dataset and hence free of data biases.
This means that given an image $x$, we
have to find and disentangle:
(i) $y$, which specifies
the dataset, and (ii) the data-unbiased content $z$, such that both of them
\emph{together} fully describe the input $x$. We
interpret this task as a minimization of their mutual information $I(z,y)$, and,
following \cite{rombach2020network}, minimize an upper bound on
it
where we model the distribution $p(z \vert y)$ with a conditional INN $\tau$
and use a standard normal prior $q(z)$.
An image $x$ from dataset $y$ is then projected onto a different dataset $y^*$
by inferring its content $z = \tau(x \vert y)$ and recombining it as $x^* =
\tau^{-1}(z \vert y^*)$.
We follow
\cite{rombach2020network} and use an autoencoder to obtain the data
representation $x$, which, once trained, allows for efficient training of the
conditional INN.
\secondpagefigure
\paragraph{On the Progress of Generative Models}
\enlargethispage{10\baselineskip}
Bad intentions behind carefully curated datasets should be considered an
exception that must not overshadow the potential of data driven systems. But
even in good-spirited attempts to improve the performance of models, it must be
understood that the training dataset is a confounding factor when assessing
effectiveness of models. While previous works which introduced higher-quality
datasets for generative modeling
controlled for this factor
\cite{karras2018progressive,karras2019stylebased,choi2020stargan}, the
perceived improvements in model quality is to a large extent influenced by
qualitative results such as Fig.~\ref{fig:ganprogress} from \cite{ganprogress}.
Our approach directly visualizes the effect of the dataset on a single model
and thereby gives an impression to what extent samples from a fixed generative
model can be improved by biasing it to datasets of different quality.
The left part of Fig.~\ref{fig:progress} shows how the
same latent code is interpreted in three different datasets corresponding to
the datasets used in the three rightmost images of Fig.~\ref{fig:ganprogress}: From
left-to-right, CelebA \cite{liu2015faceattributes}, CelebA-HQ \cite{karras2018progressive} and FFHQ
\cite{karras2019stylebased}. The transition from CelebA to CelebA-HQ shows a drastic
improvement in resolution and high-frequency detail while maintaining
person-identity well. These changes reflect the fact that CelebA-HQ uses a
``high-quality'' subset of upsampled CelebA images.
Going from CelebA-HQ to FFHQ we observe a further increase in realism through
more realistic and varied lighting and specular reflections. This might be due
to the fact that for FFHQ high quality images were collected from the
beginning.
A similar discussion applies to the right of Fig.~\ref{fig:progress} which
demonstrates even more drastic improvements when transitioning from the
AnimalFaces \cite{liu2019few} dataset to its corresponding high-quality
version, AnimalFacesHQ \cite{choi2020stargan}.

\vspace{-0.8em}
\paragraph{On the Bias of Generative Models}
While the previous paragraph discussed the implicit conception of progress that
arises when introducing new high-quality datasets alongside new models, our
method can also be used as a probing tool for biases \emph{within}
different datasets. Recently, the publication of the PULSE model
\cite{menon2020pulse} led to controversial discussions about the role of data bias
in ML \cite{lecunbias, kilcherdrama, kurenkov2020lessons}, revealing the need for
demonstrations of these data biases. Since our method can infer a
content code $z = \tau(x\vert y)$ from the data and then render it conditioned
on the dataset of interest, we are able to project a given input onto a
given
training dataset.
In Fig.~\ref{fig:secondpage}, col.~3, and Fig~\ref{fig:celebifyinputs} we apply
this method to project inputs from the FFHQ dataset onto the CelebA-HQ dataset.
Fig.~\ref{fig:celebifysamples},~\ref{fig:triplesamples} show
sampled codes $z \sim q(z) = \mathcal{N}(z \vert 0, 1)$ which are %
projected onto CelebA, CelebA-HQ and FFHQ. Besides the already discussed
difference in image quality, biases between the datasets containing images of
celebrities (CelebA, CelebA-HQ) and FFHQ, which contains a wider variety of
face images become apparent. For the FFHQ $\to$ CelebA(-HQ) direction, the most
obvious differences include a loss of diversity in facial features
such as hair style or skin appearance, a strong bias towards more make-up and
less skin wrinkles and a decreased diversity of age.
By demonstrating how societal biases
of
curated datasets are mapped into and replicated by
ML models,
awareness for the topic can be raised.

\vspace{-0.8em}
\paragraph{On Creative Applications}
The ability to visualize the \emph{same
content} viewed under differently biased datasets immediately enables
applications in creative content creation by producing visual analogies between
diverse datasets such as photographs, portaits and anime characters. We show
examples in Fig.~\ref{fig:secondpage}, col.~1, 2, Fig.~\ref{fig:portraits},
and Fig.~\ref{fig:anime}.
Thus, it can be stated that biases in data sets not only have disadvantages,
but can also be used explicitly and consciously for creative content creation
-- although one should always keep in mind that \emph{"nothing good ever comes
from face datasets"} \cite{facesuseless}.

\newpage
\section*{Ethical Implications}
\vspace{-0.5em}
\newcommand{\myspace}{\vspace{0.2em}}
Biases in Machine Learning are a sensitive issue that should not be dismissed
as being caused by the underlying data alone. Datasets are \emph{not} the sole
cause of societal bias in ML models \cite{Mitchell_2019}. Therefore, we give
some additional remarks on our presented approach.\myspace\\
  $\bullet$ \parbox[t]{0.98\linewidth}{We can compare dataset biases here since we first train an autoencoder
    on all relevant datasets \emph{combined}, i.e.\ we create one big
    dataset containing all the sub-datasets and train a single autoencoder on
    it. While the combined dataset contains yet other biases, we can faithfully
    compare biases present within the different sub-datasets.\myspace}\\
  $\bullet$ \parbox[t]{0.98\linewidth}{We then train the cINNs through maximum likelihood learning. Compared to
    GANs, this approach does not suffer from data regions being ignored
    (so-called ``mode-collapse''), but, by trying to cover all data regions,
    can suffer from estimating a ``broader'' distribution than the original
    one, i.e.\ assign too high density to regions between data points. This
    can result in a bias towards averaged data points, which, in the case of
    aligned face images, might be perceived as more attractive faces
    \cite{langlois1990}.\myspace}\\
  $\bullet$ \parbox[t]{0.98\linewidth}{We do not make a statement on how to generally investigate and solve
    societal biases of ML models. We do however provide a method to project
    images onto a given dataset by disentangling the effects of different
    datasets with respect to a generative model.\myspace}\\
  $\bullet$ \parbox[t]{0.98\linewidth}{Finally, since we use an autoencoder to obtain an efficient data
    representation, the reconstruction of an input image may not be accurate.
    This effect might be enhanced by our use of a patch-based discriminator,
    which boosts realism of the reconstructions but, to some extent, introduces
    GAN-related issues. Therefore, carefully balancing these different factors
    is important when using the proposed method.\myspace}

\vspace{-0.5em}

{\small
\bibliographystyle{IEEEtran}
\bibliography{ms}
}

\newpage
\appendix
\section{Supplementary}
\demo
\celebifyinputs
\portraits
\anime
\animalsamples
\triplesamples
\portraitsamples
\animesamples
\celebifysamples
\faceliftsamples
\figganprogress
\end{document}